\documentclass[conference]{IEEEtran}
\IEEEoverridecommandlockouts
\usepackage{cite}
\usepackage{amsmath,amssymb,amsfonts}
\usepackage{algorithmic}
\usepackage{multirow}
\usepackage{todonotes}
\usepackage{graphicx}
\usepackage{textcomp}
\usepackage{algorithm}
\usepackage{microtype}
\usepackage{xcolor}
\usepackage{microtype}
\usepackage{hyperref}
\usepackage{subfigure}
\usepackage{booktabs} 
\usepackage{amsmath, amssymb}
\usepackage{amsthm}
\usepackage{svg}
\usepackage{ulem}
\usepackage{multirow}
\newtheorem{proposition}{Proposition}

\def\BibTeX{{\rm B\kern-.05em{\sc i\kern-.025em b}\kern-.08em
    T\kern-.1667em\lower.7ex\hbox{E}\kern-.125emX}}

\begin{document}
\title{Dual-Distilled Heterogeneous Federated Learning with Adaptive Margins for Trainable Global Prototypes}
\author{
\IEEEauthorblockN{
Fatema Siddika\IEEEauthorrefmark{1},
Md Anwar Hossen\IEEEauthorrefmark{1},
Wensheng Zhang\IEEEauthorrefmark{1},
Anuj Sharma\IEEEauthorrefmark{1},
J. Pablo Mu\~noz\IEEEauthorrefmark{2},
Ali Jannesari\IEEEauthorrefmark{1}
}
\IEEEauthorblockA{\IEEEauthorrefmark{1}
\textit{Department of Computer Science, Iowa State University, USA}\\
\{fatemask, manwar, wzhang, anujs, jannesar\}@iastate.edu
}
\IEEEauthorblockA{\IEEEauthorrefmark{2}
\textit{Maro Systems}\\
pablo.munoz@maro-systems.com
}
}
\maketitle
\begin{abstract}
Heterogeneous Federated Learning (HFL) has gained significant attention for its capacity to handle both model and data heterogeneity across clients. Prototype-based HFL methods emerge as a promising solution to address statistical and model heterogeneity as well as privacy challenges, paving the way for new advancements in HFL research. These methods focus on sharing class-representative prototypes among heterogeneous clients. However, aggregating these prototypes via standard weighted averaging often yields sub-optimal global knowledge. Specifically, the averaging approach induces a shrinking of the aggregated prototypes' decision margins, thereby degrading model performance in scenarios with model heterogeneity and non-IID data distributions. We propose FedProtoKD in a Heterogeneous Federated Learning setting, leveraging clients' logits and prototype feature representations to improve system performance via an enhanced dual-knowledge distillation mechanism. The proposed framework aims to resolve the prototype margin-shrinking problem using a contrastive learning-based trainable prototype server by leveraging a class-wise adaptive prototype margin. Furthermore, the framework assesses the importance of public samples by measuring the closeness of each sample’s prototype to its class representative, thereby enhancing learning performance. FedProtoKD improved test accuracy by an average of 1.13$\%$ and up to 34.13$\%$ across various settings, significantly outperforming existing state-of-the-art HFL methods.
\end{abstract}
\begin{IEEEkeywords}
Heterogeneous FL, Dual Knowledge Distillation, Margin Shrinking, Adaptive Prototype Margin, Contrastive Learning.
\end{IEEEkeywords}
\section{Introduction}
Federated learning (FL) represents a significant breakthrough in machine learning, enabling decentralized model training while protecting data privacy. However, traditional FL methods such as FedAvg \cite{mcmahan2017communication} suffer from performance degradation due to statistical heterogeneity \cite{t2020personalized}. To address this issue, personalized FL approaches were developed to learn personalized model parameters for each client. However, most of these methods assume that all clients use the same model architecture and communicate their model updates to the server for training a shared global model\cite{zhang2023gpfl}. This approach not only leads to high communication costs \cite{zhuang2023foundation}  but also exposes client models, raising concerns about privacy and intellectual property \cite{li2021survey}. 

To address these challenges, Heterogeneous Federated Learning (HFL) \cite{tan2022fedproto} has emerged as an innovative FL paradigm that allows clients to use diverse model architectures and handle heterogeneous data without sharing private model parameters. Instead, clients exchange local knowledge, which reduces communication overhead and enhances model performance. For example, federated ensemble distillation \cite{lin2020ensemble} uses public proxy samples to replicate local outputs, effectively handling heterogeneity by removing structural constraints. However, these methods depend critically on the availability and quality of the global dataset \cite{zhang2023towards}. Data-free knowledge distillation (KD) approaches employ auxiliary models as global knowledge \cite{zhang2022fine,wu2022communication}, but the communication overhead of transmitting these models remains substantial. 

Prototype-based HFL methods~\cite{tan2022fedproto} propose sharing lightweight feature representatives, known as prototypes, as a form of global knowledge, thereby significantly reducing communication overhead. However, despite their advancements, these approaches face significant challenges when aggregating client prototypes via weighted averaging. Due to diverse model architectures and non-IID data, client prototypes often exhibit varying scales, distinct separation margins, and spatial misalignments. The prototype margin is defined as the minimum Euclidean distance separating distinct class prototypes, which is critical for classifier performance. When heterogeneous prototypes are averaged, the global prototypes suffer from a reduction in these margins. This phenomenon, known as ``Prototype Margin Shrink,'' diminishes the distinctiveness of class boundaries and degrades the quality of the global model~\cite{zhang2023semantic}. FedPKD~\cite{lyu2023prototype} uses public proxy samples to share logits, where the server filters high-quality samples via class-wise prototypes and discards the rest. Aggregating based on local sample sizes induces prototype margin shrinking, sample bias, and privacy risks from sharing data counts. Similarly, Fed2PKD~\cite{xie2024fed2pkd} aligns embeddings via prototypical distillation but restricts diverse ResNet50-152 models to a fixed 2048-dimensional layer. This constraint overlooks varying layer sizes, limiting the ability to address true model heterogeneity.

To address these limitations, this paper proposes \textbf{FedProtoKD}, a novel HFL framework that leverages enhanced knowledge distillation to mitigate prototype margin shrinkage. By transcending the uniform-dimension constraints of prior studies, the proposed method addresses true model heterogeneity. Specifically, distinct model backbones with feature embedding sizes ranging from 512 to 2048 dimensions are integrated to simulate a realistic heterogeneous environment. To bridge these disparate feature spaces, a learnable projection layer is introduced to align heterogeneous dimensions into a unified representation, enabling effective aggregation across fundamentally incompatible model structures. Furthermore, the \textbf{A}daptive \textbf{C}lass-Wise Margin-Based \textbf{T}rainable \textbf{P}rototype (ACTP) framework is designed, which employs a contrastive learning-based generator to synthesize the server's global prototypes. This trainable prototype structure is optimized via an adaptive class-wise margin that maintains significant gaps between class features, thereby enhancing class separability and reinforcing the discriminative boundaries between classes. To further enhance the reliability of the global model, a sample prioritization mechanism is implemented using the unlabeled proxy public dataset. By assigning higher importance to samples proximal to the aggregated prototypes, alignment is improved while the influence of noisy, distant data is suppressed. Additionally, a variance-based logit aggregation strategy is adopted to replace raw sample counts, thereby preserving client privacy.

Extensive experiments conducted across diverse non-IID settings demonstrate that the proposed FedProtoKD yields significant improvements in both learning performance and communication efficiency for client and server models. The main contributions are summarized as follows:
\begin{itemize}
    \item The proposed FedProtoKD framework addresses heterogeneity by utilizing learnable projection layers and a novel dual-knowledge distillation mechanism.
    \item To mitigate prototype margin shrinking, the introduction of trainable server prototypes optimized via contrastive learning with adaptive class-wise margins ensures robust class separability. 
    \item The enhancement of global learning by assessing the importance of unlabeled public proxy samples based on geometric proximity to class representatives, prioritizing high-quality samples for distillation.  
    \item The execution of extensive experiments across heterogeneous models and datasets demonstrates that FedProtoKD outperforms state-of-the-art methods with server accuracy improvements ranging from 1.13\% to 34.13\%. These results confirm that the class-wise adaptive margin enhances prototype separation.
\end{itemize}
\section{Problem Formulation and Motivation}
In Heterogeneous Federated Learning (HFL), diverse client architectures create dimensional mismatches that make direct weighted aggregation infeasible, shifting collaboration toward knowledge exchange rather than parameter sharing. Logits enable cross-model knowledge transfer through public data, while prototypes capture class-wise feature representations from private data. This section also discusses the limitations of relying solely on classification learning and highlights the prototype margin-shrinking problem in feature learning, which motivates our adaptive solution.
\begin{figure}[htbp]
\centerline{\includegraphics[width=0.8\linewidth]{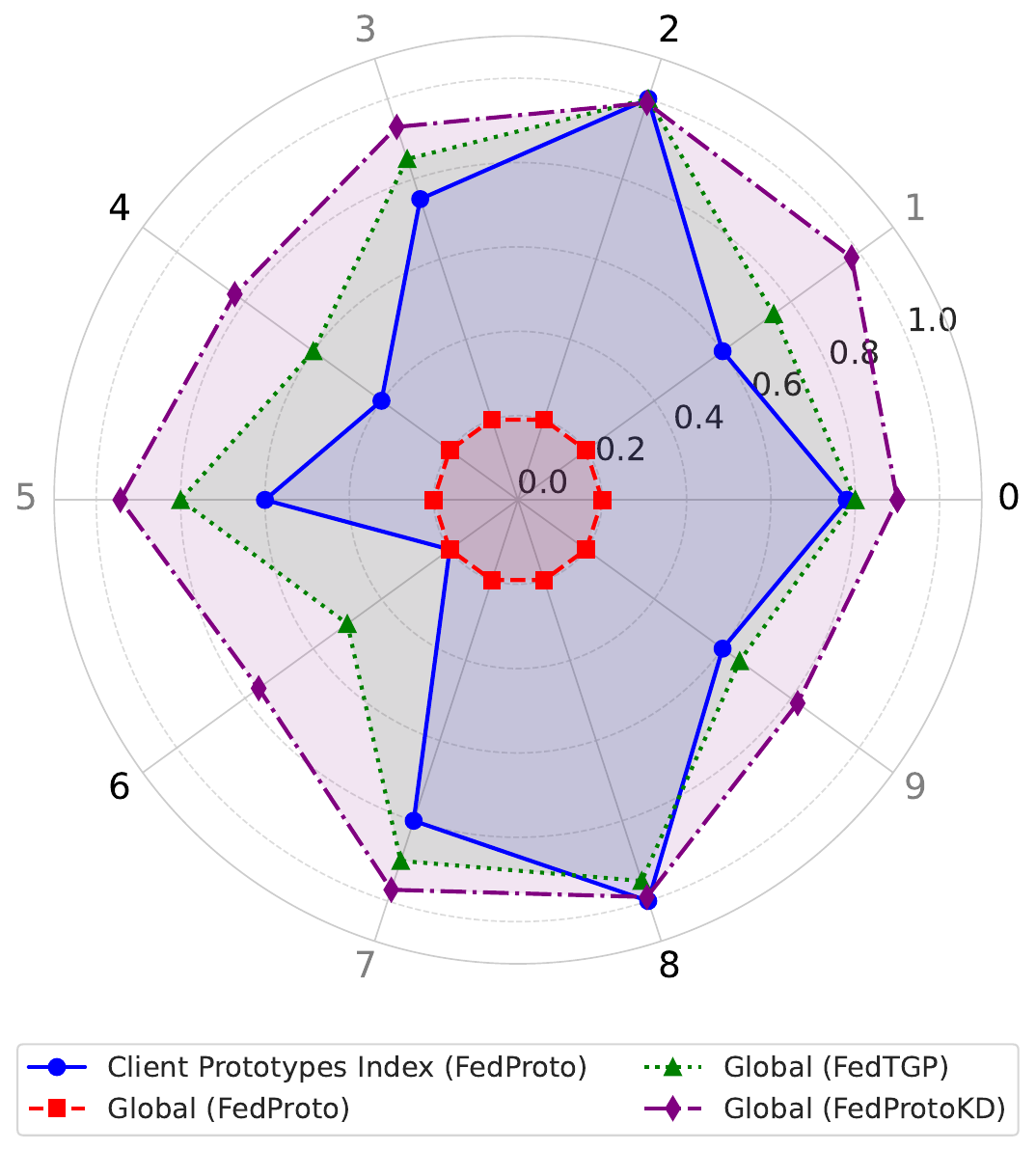}}
\caption{Illustrating normalized prototype margins across CIFAR-10 classes (0–9), where the concentric grid lines (0.0 to 1.0) represent the magnitude of the normalized margin. The prototype margin is defined as the minimum Euclidean distance separating distinct class prototypes, while the maximum margin denotes the highest value observed across all clients per class. The blue line represents the maximum margin observed among all local clients and, the red line depicts the aggregated prototype margin of FedProto, indicating the baseline margin shrinkage that occurs during standard aggregation. FedProtoKD generates global prototypes with enhanced inter-class separability, thereby maximizing geometric distinguishability and mitigating the shrinkage problem.}
\label{fig:margin_shrinking}
\end{figure}
\begin{figure*}[htbp]
\centerline{\includegraphics[width=1\linewidth]{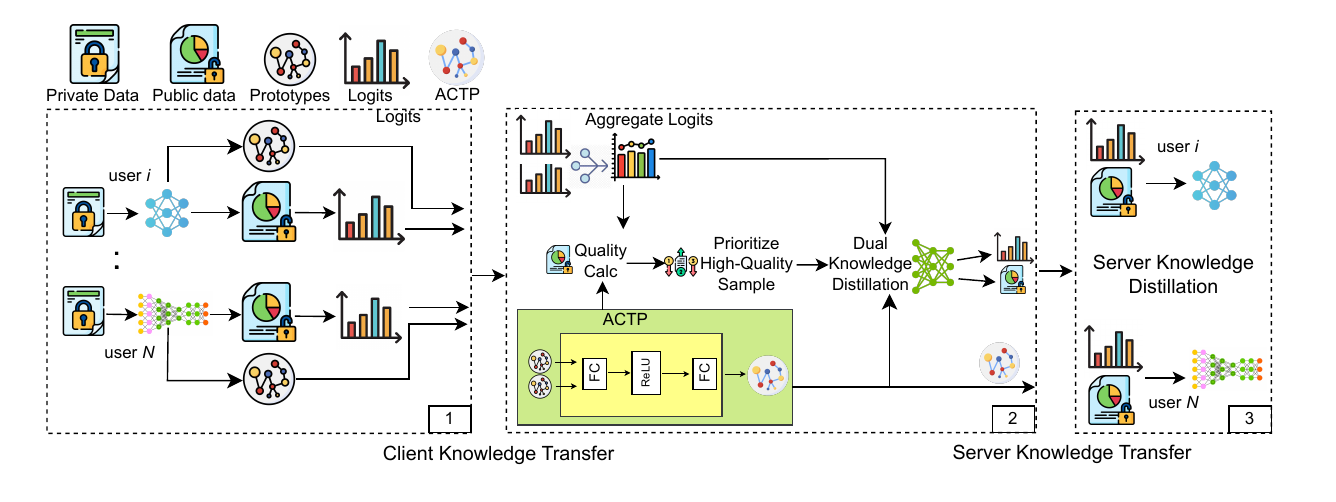}}
\caption{Overview of the FedProtoKD framework. \textbf{(1) Client Knowledge Transfer:} Clients compute local prototypes and logits using unlabeled public data and transmit them to the server. \textbf{(2) Server Aggregation:} The server employs a dual-branch strategy: the ACTP module synthesizes adaptive global prototypes via a non-linear generator, while the logit branch prioritizes high-quality samples based on prediction variance. These components are fused via Dual Knowledge Distillation. \textbf{(3) Server Knowledge Transfer:} The refined global knowledge is distilled back to clients to align local feature representations.}
\label{fig:system_diagram}
\end{figure*}
\subsection{Prototype Margin}
A prototype serves as a compact representation of a class, formally defined as the centroid of the feature distribution. For client $k$, the prototype $P_k^c$ is computed by averaging the feature vectors of all local samples belonging to class $c$ in the set $D_k^c$.
\begin{equation}
    P_k^c = \frac{1}{|D_k^c|} \sum_{x \in D_k^c} \mathcal{F}_{\theta_k}(x),
\end{equation}
where, $\mathcal{F}_{\theta_k}(\cdot)$ denotes the feature extractor. To quantify inter-class separability, FedProto \cite{tan2022fedproto} defines the prototype margin as the minimum Euclidean distance between the prototype of class $c$ and its nearest neighbor: $\delta^c = \min_{j \neq c} \| P^c - P^j \|_2$. Accordingly, the maximum margin, $\Delta_{max} = \max_{c} (\delta^c)$, denotes the largest among all class prototype margins, reflecting the highest inter-class separability achieved by the system, as shown in Figure \ref{fig:margin_shrinking}. Maximizing $\Delta_{max}$ promotes the separation of class feature distributions and enhances discriminative capability.
\subsection{Prototype Dimensionality in Heterogeneous Models}
In heterogeneous federated learning, clients employ diverse model architectures resulting in prototype representations with incompatible dimensions. This structural inconsistency renders direct aggregation impossible. While centralized methods like DINO~\cite{caron2021emerging} and DeiT~\cite{touvron2021training} use projection heads to reduce dimensionality, they do not address aligning prototype representations across distinct architectures. Bridging these dimensional discrepancies is essential to enable global aggregation without discarding critical semantic information or compromising local inductive biases.
\subsection{Aggregated Prototypes' Margin Shrinking Problem}
\begin{figure}[htbp]
\centerline{\includegraphics[width=.7\linewidth]{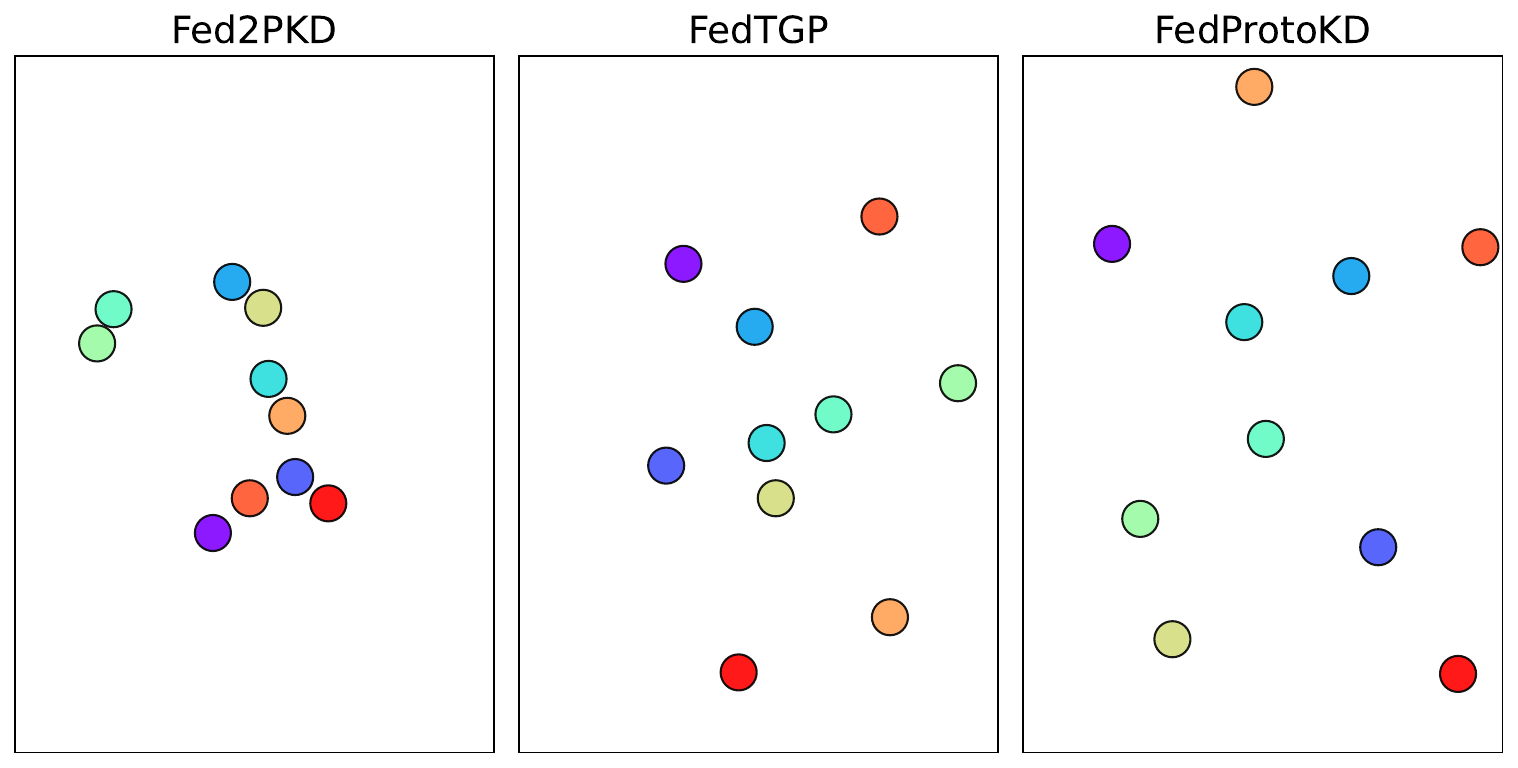}}
\caption{t-SNE visualization of local class prototypes after integrating global prototype updates, where each point represents the feature centroid learned by a client for a specific class. While Fed2PKD and FedTGP exhibit crowded clusters indicative of margin shrinkage, FedProtoKD maintains distinct separation. This confirms that the ACTP mechanism effectively enforces geometric separability, preventing margin collapse.}
\label{fig:tSNE_class_mean_features_points}
\end{figure}

Client prototypes vary significantly in scale and distribution due to data and model heterogeneity. As a result, the standard weighted aggregation $\sum w_k P_k^c$ employed in FedProto \cite{tan2022fedproto} blurs class boundaries when applied to these divergent features. This diminishes inter-class separability and the prototype margin $\delta^c$. This contraction of the global maximum margin $\Delta_{max}$ is defined as \textit{Prototype Margin Shrink}. To mitigate prototype margin shrink, FedTGP \cite{JZhang2024} introduces trainable global prototypes optimized via an adaptive margin. This approach aims to enforce clearer decision boundaries by explicitly separating class centers. However, applying a uniform margin magnitude across all classes ignores class-specific separability. As illustrated in Figure \ref{fig:tSNE_class_mean_features_points}, this rigid constraint forces unnecessary separation on compact classes. Hence, the method suffers from over-regularization, distorting the feature space and disrupting natural class relationships, as shown in Figure \ref{fig:margin_shrinking}.
To overcome the margin-shrinking problem, this paper proposed a novel framework that incorporates an Adaptive Class-Wise Margin-Based $\xi^c(t)$ Trainable Prototype (ACTP), utilizing a contrastive learning-based generator to generate a trainable server prototype. ACTP dynamically adjusts class-specific margin throughout the training process and penalizes each class prototype individually based on its specific distribution characteristics. This class-specific calibration prevents over-regularization for tightly clustered classes while ensuring robust separation for dispersed ones. The figure \ref{fig:tSNE_class_mean_features_points} illustrates the feature separation, demonstrating that this strategy optimizes the generation of class prototypes, which ensures superior feature alignment and significantly improves overall model performance.
\subsection{Quality-Diversity Trade-off in Public Data Logits}
In distillation-based federated learning, client logits on public data serve as a proxy for local knowledge, effectively transferring learned classification patterns to the server. Prior work, FedMD \cite{FedMD2019} and Fed2PKD \cite{xie2024fed2pkd}, treats these logits equally, failing to account for the varying reliability of heterogeneous clients. Conversely, FedPKD \cite{lyu2023prototype} attempts to mitigate noise by filtering samples based on feature distance. While excluding samples reduces noise, it inevitably lowers data diversity and increases the risk of overfitting to a specific subset of the distribution. To address these limitations, importance weights are quantified and assigned to all public samples, ensuring effective distillation from high-quality logits while preserving the robustness provided by the diversity of the public dataset.
\section{Methodology}
\subsection{Client Learning and Dual Knowledge Sharing}
The framework consists of a Federated Learning (FL) framework with $K$ participating clients tackling a classification task with $C$ classes. Algorithm \ref{algo:fedprotokd_flow} summarizes the complete execution flow of FedProtoKD. Each client $k \in \{1, \dots, K\}$ possesses a local private dataset $D_k = \{(x_i, y_i)\}$, where the local model $\mathcal{F}_{\theta_k}$ is trained by optimizing the cross-entropy loss $\mathcal{L}_{\text{CE}}$:
\begin{equation}
\label{eq:local_train_t_round}
\min_{\theta_k} \sum_{(x_i, y_i) \in D_k} \mathcal{L}_{\text{CE}} \left(\mathcal{F}_{\theta_k}(x_i), y_i \right).
\end{equation}
\subsubsection*{Prototype Feature Alignment via Learnable Projection}
To address the channel inconsistencies arising from heterogeneous models, a learnable projection layer $h_{\phi_k}(\cdot)$ incorporate directly into the local training process. This layer maps diverse feature representations to a unified channel dimension $d$. Formally, let $F_k(x)$ denote the local feature map extracted by the model for a sample $x$. The aligned feature map $F'_k(x)$ is computed and subsequently flattened to yield the final feature vector $z_k(x)$:
\begin{equation}
\label{eq:local_class_wise_proto_calc}
    F'_k(x) = h_{\phi_k}(F_k(x)), \quad z_k(x) = \text{vec}(F'_k(x)),
\end{equation}
Here, $\text{vec}(\cdot)$ reshapes the tensor by collapsing the spatial dimensions into a single vector. This approach preserves the spatial structural information within the aligned feature vector $z_k(x)$, ensuring that the computed prototypes $P_k^c$ capture both the semantic channel features and their spatial arrangement.
\subsubsection*{Dual Knowledge Sharing}
Following local optimization, these flattened feature vectors $z_k(x)$ are utilized to compute the class-wise prototypes $P_k^c$ by averaging the vectors for all samples of class $c$ in the local dataset $D_k$. Simultaneously, clients perform inference on the unlabeled public dataset $D_p$ to extract soft logits. Finally, each client shares these aligned class-wise prototypes and public logits with the server, enabling dual knowledge transfer.
\subsection{Adaptive Class-Wise Margin-Based Trainable Prototype}
The server learns from prototypes and logits to distill a global model. However, in heterogeneous settings, weighted prototype averaging induces margin shrinkage due to the high variance in client feature distributions. To overcome this, the server employs the Adaptive Class-wise Margin-based Trainable Prototype (ACTP) framework to synthesize prototypes via a contrastive learning objective, dynamically adjusting margins to reflect the intrinsic geometry of class distributions.
\subsubsection*{Design of Contrastive Learning Based Generator}
A generative framework synthesizes refined server prototypes $\tilde{P}^c$ from latent class seeds. A trainable vector $\hat{P}^c \in \mathbb{R}^d$ is initialized for each class $c \in C$, serving as a latent anchor representing the semantic and spatial structure of that class. These vectors are processed by a server-side network to produce global prototypes, computed as $\tilde{P}^c = \mathcal{G}_{\theta}(\hat{P}^c; \theta)$. The generator $\mathcal{G}_{\theta}$ transforms latent seeds $\hat{P}^c$ into the flattened feature space via:
\begin{equation}
    \tilde{P}^c = \mathcal{G}_{\theta}(\hat{P}^c) = W_2 \cdot \sigma(W_1 \hat{P}^c + b_1) + b_2,
\end{equation}
where $W_1, W_2$ are weight matrices mapping to $\mathbb{R}^d$, $b_1, b_2$ are biases, and $\sigma(\cdot)$ is the ReLU function. This introduces necessary non-linearity to capture complex class boundaries that linear mappings fail to model. Parameters $\theta = \{W_1, W_2, b_1, b_2\}$ are shared across classes, enforcing a universal rule that prevents memorizing class-specific noise while maintaining geometric consistency.
The generator $\mathcal{G}_{\theta}$ is computationally efficient, operating on low-dimensional seeds ($d=512$) rather than raw data. Its lightweight MLP architecture has $\mathcal{O}(d^2)$ time complexity per prototype. Since training uses only aggregated prototypes where $\tilde{P}^c \ll P^C_K$, the total optimization cost $\mathcal{O}(E \cdot \tilde{P}^c \cdot d^2)$ remains negligible compared to client-side training.
\begin{algorithm}
\caption{FedProtoKD}
\begin{algorithmic} 
\label{algo:fedprotokd_flow}
\STATE \textbf{Input:} Client $k \in \{K\}$, Public data $D_p$, private data $D_k$, Server model $\mathcal{F}_{\theta_G}$, Client model $\mathcal{F}_{\theta_k}, k \in \{K\}$, Rounds $T$
\STATE \textbf{Output:} Server $\mathcal{F}_{\theta_G}$, Client $\mathcal{F}_{\theta_k}, k \in \{K\}$
\FOR{$t = 0, 1, \dots, T-1$}
    \FOR{each client $k$ in parallel}
        \STATE $t = 0, \theta_k^t \gets$ Local Training on $D_k^t$, Eq. \eqref{eq:local_train_t_round}
        \STATE $t > 0, \theta_k^t \gets$ Local Training on $D_k^t$, Eq. \eqref{eq:local_train_t+1_round}
        \STATE Class wise avg clients prototype, $P^{c,t}$, Eq. \eqref{eq:local_class_wise_proto_calc}
        \STATE Send logits $\mathcal{F}_{\theta_k}^t(x_i)$ and prototype $P^{c,t}$ to server
        \vspace{2pt}
    \ENDFOR 

    \STATE \text{Server aggregate clients logits}, $\hat{L}^t$, Eq. \eqref{eq:server_logits_aggregation}
    
    \STATE Server generate $\tilde{P}^{c,t}$ using ACTP,  Eq. \eqref{eq:class_wise_adaptive_margin_proposed}

    \STATE Prototype \& logits-based sample importance, Eq.\eqref{eq:public_data_filtering_part2}
    
    \STATE $\theta^t_G \gets $ KD to Server model $\mathcal{F}(\theta^t_G)$ on $D_p$, Eq. \eqref{eq:server_training}
    
    \STATE Share server models' logits $\mathcal{F}^t_{\theta_G}(x_i)$ \& prototype $\tilde{P}^{c,t}$ to clients
   
    \FOR{\textbf{each} client $k$ do in parallel}
        \STATE Receive logits $\mathcal{F}^t_{\theta_G}(x_i)$ and prototype $\tilde{P}^{c,t}$
        \STATE $\theta_k^t \gets$ update client model with $\mathcal{F}^t_{\theta_G}(x_i)$, Eq. \eqref{eq:local_m_update_with_g_KD}
    \ENDFOR
\ENDFOR
\STATE \textbf{return} $\mathcal{F}_{\theta_G}, \mathcal{F}_{\theta_k}$
\end{algorithmic}
\end{algorithm}
\subsubsection*{Adaptive Margin and Optimization Objective}
Enforcing a uniform maximum margin imposes rigidity that often over-regularizes tightly clustered classes, distorting their natural distribution. To mitigate this, an adaptive class-wise margin $\xi^c(t)$ is introduced during training round $t$. First, global class-wise prototype centers $Q_t^c$ are computed by averaging received client prototypes $P^c_k$. Subsequently, $\xi^c(t)$ is derived from the Euclidean separability of these centers. To ensure optimization stability and prevent gradient explosion, this margin is calculated as the minimum distance to the nearest distinct class cluster, margin clipped by a growth threshold $\zeta$:
\begin{equation}
\label{eq:cluster_wise_margin}
\xi^c(t) = \min_{c' \neq c} \left( \delta(Q^c_t, Q^{c'}_t), \zeta \right)
\end{equation}
where $\delta(\cdot, \cdot)$ denotes the Euclidean distance. $\xi^c(t)$ scales prototypes based on semantic similarity so that related classes remain closer to preserve structure, while distinct ones are separated. The generator minimizes a contrastive loss using these flexible, similarity-aware boundaries. This objective pulls the generated server prototype $\tilde{P}^c$ toward corresponding client prototypes $P^c_k$, ensuring robust separation from incorrect classes by at least the adaptive margin $\xi^c(t)$. The loss function is defined as:
\begin{equation}
\label{eq:class_wise_adaptive_margin_proposed}
\mathcal{L}_{P^c} = \sum_{k \in S_t} -\log  \frac{ e ^ {-\left(\Delta(P_k^c, \tilde{P}^c) + \xi^c(t)\right)}}{e^{ - \left(\Delta(P_k^c, \tilde{P}^c) + \xi^c(t)\right)} + \sum_{c' \neq c} e^{ - \Delta(P_k^c, \tilde{P}^{c'}) }},
\end{equation}
where $\Delta(P_k^c, \tilde{P}^c)$ is the Euclidean distance between the client and server prototypes.
\subsubsection*{Convergence and Stability Analysis} The boundedness of the gradient descent process is examined to guarantee that the ACTP framework remains numerically stable. 
\begin{proposition}
\textit{Boundedness of Gradients under Adaptive Clipping:} For the loss function defined in Eq. \ref{eq:class_wise_adaptive_margin_proposed}, the gradient magnitude with respect to the generated prototype $\nabla_{\tilde{P}^c} \mathcal{L}$ is strictly bounded, provided the adaptive margin $\xi^c(t)$ satisfies the constraint $\xi^c(t) \leq \zeta$.
\end{proposition}
\begin{proof} The objective function $\mathcal{L}_{P^c}$ operates as a margin-penalized Softmax cross-entropy. In conventional metric learning, unconstrained margin maximization frequently induces numerical instability; the optimizer may force the effective distance $d_{pos} = \Delta(P_k^c, \tilde{P}^c) + \xi^c(t)$ toward $-\infty$ to minimize the loss, leading to gradient explosion. However, formulation in Eq. \ref{eq:cluster_wise_margin} enforces a hard geometric constraint $\xi^c(t) \leq \zeta$. By structurally bounding the margin, the shifted distance metric is confined to a compact range relative to the negative class distances. Then, the gradient $\nabla_{\tilde{P}^c} \mathcal{L}$ satisfies the Lipschitz continuity condition with a constant $L$ dependent on $\zeta$. This constraint prevents the generator $\mathcal{G}_{\theta}$ from projecting prototypes into asymptotic regions, ensuring bounded parameter updates for $\theta$ and convergence to a stable local minimum.
\end{proof}
\begin{figure}[htbp]
    \centering
    \includegraphics[width=0.9\linewidth]{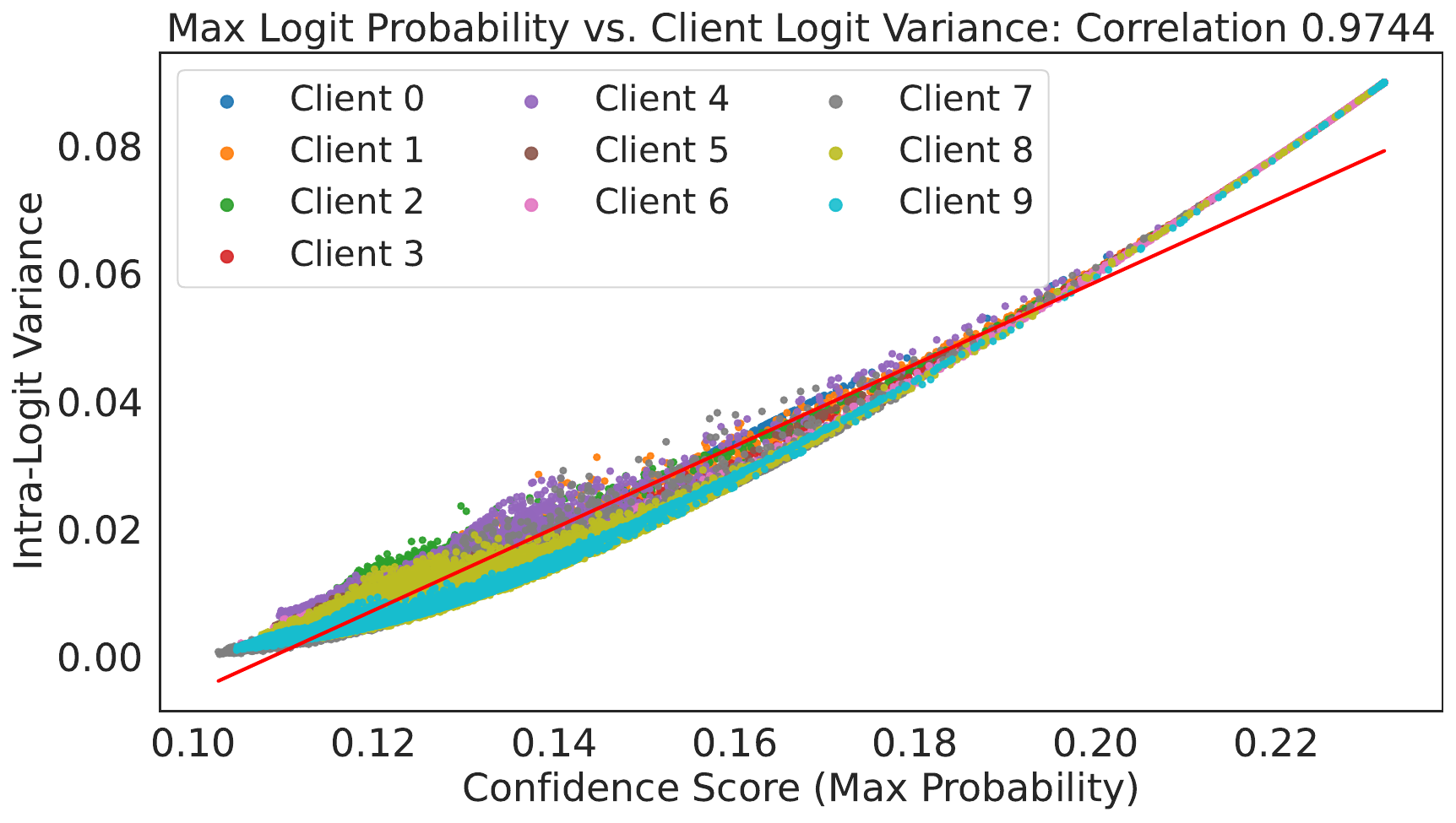}    \caption{The intra-logit variance exhibits a higher confidence score, and its correlation of 0.9744 aligns closely with the intuition behind variance-based logit aggregation.}
    \label{fig:logits_confidenceScore_vs_variance}
\end{figure}
\subsection{Variance-Weighted Global Logit Aggregation}
As clients share their logits on unlabeled public samples with the server, these predictions are aggregated to derive a global consensus. Prior to aggregation, the quality of each logit sample is evaluated, as it serves as a direct indicator of the reliability of the client's local learning. Figure \ref{fig:logits_confidenceScore_vs_variance} demonstrates that a higher variance in logit values corresponds to increased model confidence for a given sample. The variance of each logit is computed to quantify prediction quality. These variance metrics are subsequently utilized to weight the contributions of each client during the aggregation process.
The aggregated logits \( \hat{L}^t(x_i) \) for a sample \( x_i \) are computed as:
\begin{equation}
\label{eq:server_logits_aggregation}
\hat{L}^t(x_i)  = \sum_{k, x_i \in D_p} \frac{\sigma^2( \mathcal{F}_{\theta_k}^t(x_i) )}{\sum_{j \in K, x_i \in D_{p}} \sigma^2( \mathcal{F}_{\theta_j}^t(x_i) )}\mathcal{F}_{\theta_k}^t(x_i).
\end{equation}
\subsection{Quality-Aware Public Sample Prioritization}
\begin{figure}[htbp]
    \centering
    \includegraphics[width=.9\linewidth]{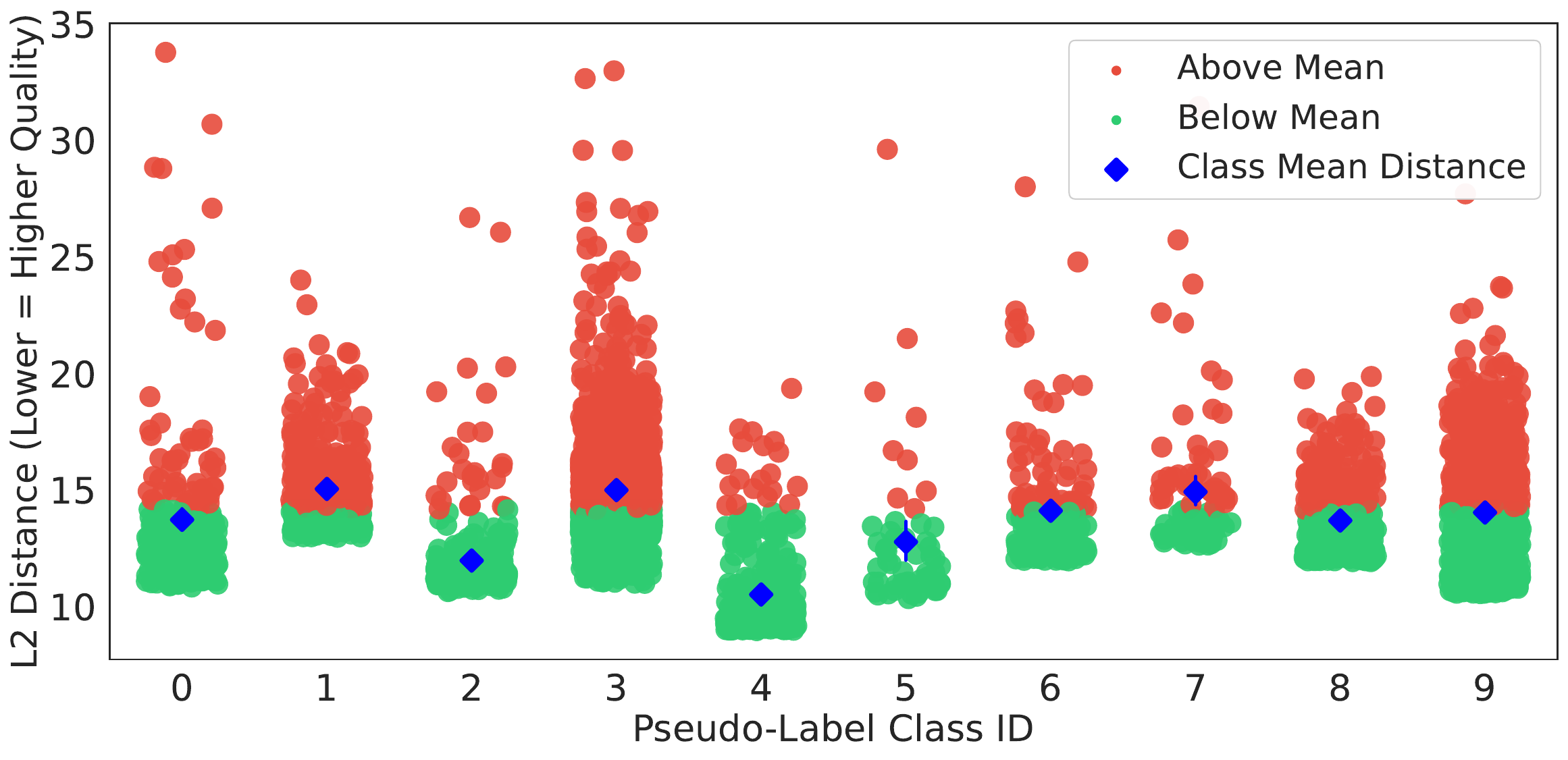}
    \caption{The plot visualizes the L2 distance between public sample feature vectors and their corresponding global class prototypes $\tilde{P}^c$. Samples are categorized relative to the class mean distance, which is marked by a blue diamond. Green points represent high-quality samples located closer to the prototype, while red points indicate lower-quality samples situated further away. The significant variance observed within classes demonstrates that samples possess unequal reliability, motivating the proposed sample-wise importance weighting mechanism $\mathcal{I}_i$ for the Global Knowledge Distillation loss.}
    \label{fig:PublicImportance}
\end{figure}

For each public sample $x_i$, the server identifies the pseudo-label class $\tilde{y}_i$ using the aggregated global logits $\hat{L}^t(x_i)$:
\begin{equation}
\tilde{y}_i = \arg\max_{y_i \in [0, N-1]} \hat{L}^t(x_i).
\end{equation}
Samples distant from their pseudo-label prototype are considered low-quality, while proximity indicates high semantic reliability. Therefore, small $L2$ distances denote high-quality contributors beneficial for global updates. Although low-quality samples are retained to preserve diversity, their influence is reduced, as illustrated in Fig. \ref{fig:PublicImportance}. To quantify this, the inverse $L2$ distance is calculated as $\tilde{d}(x_i) = 1/(d(x_i) + \epsilon)$, where $\epsilon$ prevents division by zero. Prioritization is governed by a hyperparameter $\varphi$, while a sigmoid scaling function centered on the normalized median distance $c_d$ suppresses distant samples. The final importance score $\mathcal{I}_i$ is computed as:
\begin{equation}
\label{eq:public_data_filtering_part2}
   \mathcal{I}_i = \varphi \cdot (1 + \hat{d}(x_i)) + (1 - \varphi) \cdot \left(1 - \frac{1}{1 + e^{(-k \cdot (\hat{d}(x_i) - c_d) )}}\right),
\end{equation}
where $k$ controls the sigmoid steepness to ensure smoothness, following Curriculum Learning \cite{bengio2009curriculum} principles.
\subsection{Global Model Learning via Dual Distillation}
The server trains the global model using the aggregated logits $\hat{L}^t$ and incorporates the importance factor $\mathcal{I}$ of public data samples to distill client knowledge into the server model. The generated trainable prototypes $\tilde{P}^{c,t}$ are used to enable dual knowledge distillation. The Kullback–Leibler divergence (KL) loss is computed using the aggregated logits $\hat{L}^t$, while the Cross-Entropy (CE) loss is calculated between the pseudo-label $y_i$ and the model output $\mathcal{F}^t_{\theta_G}(x_i)$. The importance factor $\mathcal{I}$ adjusts the CE term, allowing the model to focus more on informative samples during distillation while still learning from diverse public data. The knowledge distillation loss $\mathcal{L}_{kd}$ for the model is defined as:
\begin{equation}
\mathcal{L}_{\text{kd}} = \frac{1}{|D_p|} \sum_{(x_i) \in D_p} \mathcal{L}_{\text{KL}}(\hat{L}^t(x_i), \mathcal{F}^t_{\theta_G}(x_i)) 
+
\nonumber
\end{equation}
\label{eq:server_kd_loss}
\begin{equation}\frac{1}{|D_p|} \sum_{(x_i, \tilde{y}_i) \in D_p} \mathcal{L}_{\text{CE}}(\mathcal{F}^t_{\theta_G}(x_i), \tilde{y}_i) \cdot \mathcal{I}_i.
\end{equation}

Aim to minimize the distance between the feature vectors $\mathcal{V}^t_{\theta_G}$ of each public data sample $(x_i, \tilde{y}_i) \in D_p$ and their corresponding global pseudo-label prototypes $\tilde{P}^{t,\tilde{y}_i}$. The focus of the mean squared error loss $\mathcal{L}_{MSE}$ is to minimize the prototype distance and improve the alignment between the client and server models. The prototype loss term $\mathcal{L}_{pl}$ in the server model training is defined as:
\begin{equation}
\label{eq:server_mse_loss}
\mathcal{L}_{pl} = \frac{1}{|D_p|} \sum_{(x_i, \tilde{y}_i) \in D_p} \mathcal{L}_{\text{MSE}}(\mathcal{V}^t_{\theta_G}(x_i), \tilde{P}^{t,\tilde{y}_i}).
\end{equation}
The global model is optimized by solving the following objective function to minimize the loss during training:
\begin{equation}
\label{eq:server_training}
\mathcal{F}(\theta_G) = \mathcal{L}_{\text{kd}} . \Upsilon + \mathcal{L}_{\text{pl}} . (1 - \Upsilon).
\end{equation}
\subsection{Distilling Global Knowledge to Clients}
After distilling the knowledge into the server model, the server transfers the learned global knowledge to the clients, which includes the logits of the global model $\mathcal{F}_{\theta_G}^t(x_i)$ on unlabeled samples and prototypes $\tilde{P}^c$ , to update the local models.  Upon receiving the logits $\mathcal{F}_{\theta_G}^t(x_i)$, the client first generates a pseudo-label \( \tilde{y}^s_i \) for each unlabeled public sample to guide the local training:
\begin{equation}
    \tilde{y}^s_i = \arg\max_{\text{label} \in [0, N-1]} \mathcal{F}_{\theta_G}^t(x_i).
\end{equation}
To prevent the client model from overfitting due to limited local data and to enhance generalization, each client first trains using the public data subset $D_p$ and the knowledge received from the server. The optimization objective for this distillation phase is defined as:
\begin{equation}
\label{eq:local_m_update_with_g_KD}
\begin{split}
    \min_{\theta_c} \sum_{(x_i, \tilde{y}_i) \in D_p} \Big( & \gamma \mathcal{L}_{KL}(\mathcal{F}_{\theta_c}^t(x_i), \mathcal{F}_{\theta_G}^t(x_i)) \\
    & + (1-\gamma)\mathcal{L}_{CE}(\mathcal{F}_{\theta_c}^t(x_i), \tilde{y}^s_i) \Big)
\end{split}
\end{equation}
Here, $\mathcal{L}_{KL}$ aligns logits via KL Divergence and $\mathcal{L}_{CE}$ applies Cross-Entropy loss using pseudo-labels. Standard local training occurs solely in the first round via Equation \ref{eq:local_train_t_round}. In subsequent rounds $(t+1)$, clients leverage global prototypes $\tilde{P}^{c,t}$ to regularize training on private data $D_c$, narrowing the distance to global representations. Therefore, the client optimization objective is formulated as:
\begin{equation}
\label{eq:local_train_t+1_round}
\begin{split}
    \min_{\theta_c} \sum_{(x_i, y_i) \in D_c} \Big(& \mathcal{L}_{CE}(\mathcal{F}_{\theta_c}^{t+1}(x_i), y_i) \\
        & + \epsilon \mathcal{L}_{MSE}(\mathcal{V}_{\theta_c}^{t+1}(x_i), \tilde{P}^{t, y_i})\Big),   
\end{split}
\end{equation}
where $\mathcal{V}_{\theta_c}^{t+1}(x_i)$ denotes the client’s feature representation and $\tilde{P}^{t, y_i}$ is the global prototype corresponding to the ground truth class $y_i$.
\section{Experiments Design}
\subsection{Datasets and Heterogeneity Configuration}
Experiments utilize the CIFAR-10, CIFAR-100, and Tiny-ImageNet datasets. To simulate a realistic federated environment, 2,500 unlabeled samples serve as a public dataset, with the remaining data partitioned among clients under two heterogeneous settings. The practical setting employs a Dirichlet distribution, where $\alpha=0.1$ denotes extreme heterogeneity and $\alpha=0.3$ indicates moderate heterogeneity. The pathological setting \cite{mcmahan2017communication} assigns a fixed class count $k$ per client. Extreme heterogeneity sets $k=3$ for CIFAR-10 and $k=20$ for CIFAR-100, while moderate heterogeneity increases these to $k=5$ and $k=40$. Results represent the mean of two runs using identical seeds and hyperparameters to ensure fair and consistency.
\subsection{Backbone Architectures}
A diverse suite of backbone architectures is employed for system heterogeneity. This includes ResNet18 and ResNet34 with 512-dimensional layers, alongside ResNet50, ResNet101, and ResNet152 featuring 2048-dimensional layers. Additionally, GoogleNet with 1024 dimensions and MobileNet-v2 with 1280 dimensions are utilized. ResNet34 consistently serves as global server model for aggregation and as a homogeneous model.
\begin{figure}[htbp]
\centerline{\includegraphics[width=0.95\linewidth]{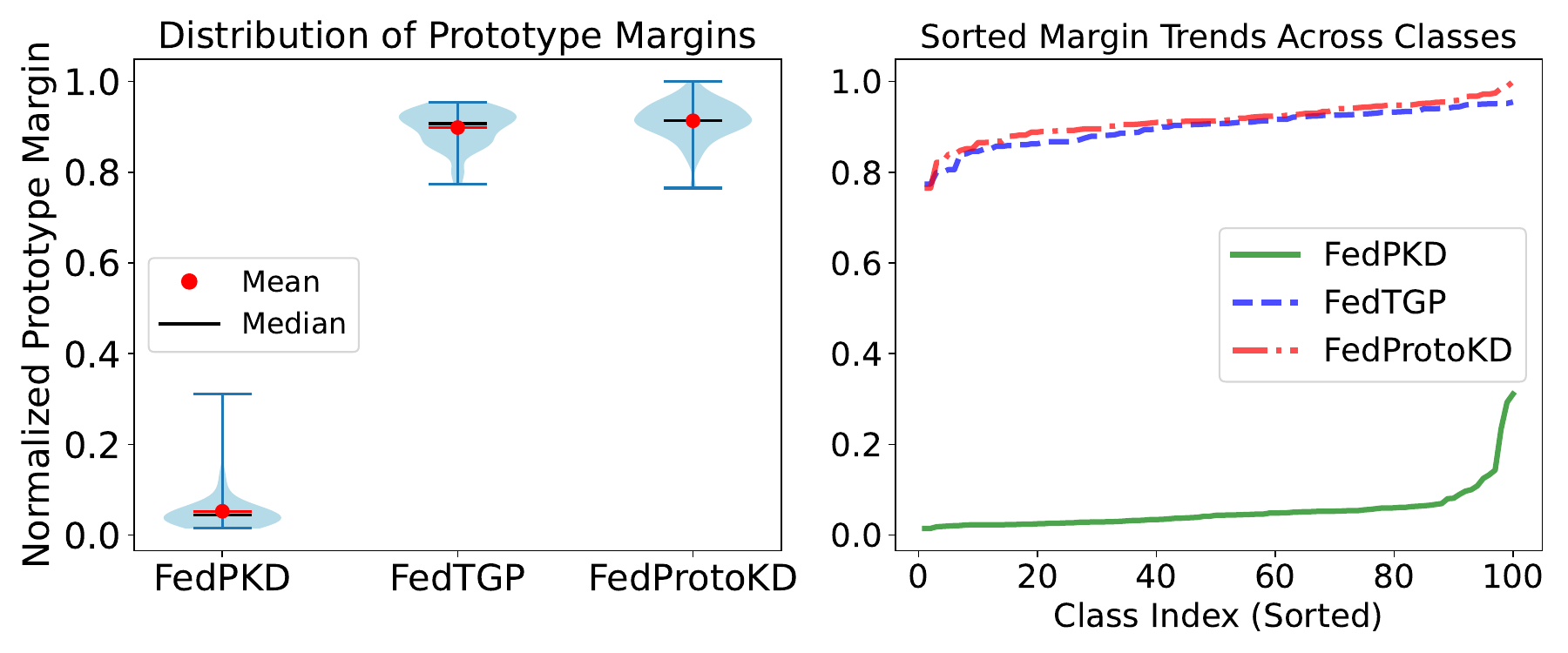}}
\caption{Distribution and Trends of Class Prototype Margins in CIFAR-100}
\label{fig:margin_shrink_solve}
\end{figure}
\subsection{Baseline Methods}
The proposed FedProtoKD framework is benchmarked against a comprehensive set of state-of-the-art federated learning approaches to validate its efficacy. Baselines prototype-based distillation-based methods such as FedProto \cite{lin2020ensemble}, FedPKD \cite{lyu2023prototype} and FedTGP \cite{JZhang2024}, as well as FedProx \cite{TLiFedProx2020}, Fed2PKD \cite{xie2024fed2pkd}. Additionally, an ablated variant, FedProtoKD-$\zeta$, is evaluated; this variant applies an adaptive maximum margin to all trainable class prototypes in the server while utilizing the full set of public samples.
\begin{table*}[htbp]
\centering
\caption{Server Test Accuracy (\%) under Extreme and Moderate Heterogeneity on Heterogeneous Models}
\label{table:combined_server_test_accuracy_hetero_model}
\begin{tabular}{|c|cc|cc|cc|cc|}
\hline
\multirow{3}{*}{} & \multicolumn{4}{c|}{Extreme Data Heterogeneity} & \multicolumn{4}{c|}{Moderate Data Heterogeneity} \\ \cline{2-9}
 & \multicolumn{2}{c|}{$\alpha=0.1$} & \multicolumn{2}{c|}{$k = 3/20$} & \multicolumn{2}{c|}{$\alpha=0.3$} & \multicolumn{2}{c|}{$k = 5/40$} \\ \cline{2-9}
Methods & CIFAR-10 & CIFAR-100 & CIFAR-10 & CIFAR-100 & CIFAR-10 & CIFAR-100 & CIFAR-10 & CIFAR-100  \\ \hline
Fed2PKD & 54.84& 24.98& 27.81& 15.70& 42.53&  31.20& 55.47& 19.10\\
FedPKD & 62.58& 29.11 & 60.58 & 29.69 & 68.52 & 30.04 & 70.12& 29.23\\ 
FedProtoKD-$\zeta$ & 62.11& 30.23 & 60.61 & 29.05 & 68.61 & 31.76 & 69.50& 30.15\\ \hline
\textbf{FedProtoKD} & \textbf{63.49}& \textbf{31.80}& \textbf{61.97}& \textbf{30.88}& \textbf{69.45}& \textbf{32.63}& \textbf{71.31}& \textbf{30.29}\\ \hline
\end{tabular}
\end{table*}
\begin{table*}[htbp]
\centering
\caption{Clients Test Accuracy (\%) under Extreme and Moderate Heterogeneity on Heterogeneous Models}
\label{table:combined_client_test_accuracy_hetero_model}
\begin{tabular}{|c|cc|cc|cc|cc|}
\hline
\multirow{3}{*}{} & \multicolumn{4}{c|}{Extreme Data Heterogeneity} & \multicolumn{4}{c|}{Moderate Data Heterogeneity} \\ \cline{2-9}
 & \multicolumn{2}{c|}{$\alpha=0.1$} & \multicolumn{2}{c|}{$k = 3/20$} & \multicolumn{2}{c|}{$\alpha=0.3$} & \multicolumn{2}{c|}{$k = 5/40$} \\ \cline{2-9}
Methods & CIFAR-10 & CIFAR-100 & CIFAR-10 & CIFAR-100 & CIFAR-10 & CIFAR-100 & CIFAR-10 & CIFAR-100  \\ \hline
FedProto & 87.16&  59.98& 89.38& 59.13& 79.51&  44.22& 82.54&60.31\\
FedTGP & 87.32& 60.42 &  91.95&  60.03& 80.58& 44.74 &  83.33&  60.65\\ 
Fed2PKD & 78.95& 47.72 & 76.37& 48.90& 71.57& 35.30& 78.48& 43.85\\
FedPKD & 85.27 & 59.92 & 90.05& 60.27 & 79.52 & 42.04 & 83.91 & 59.80 \\ 
    FedProtoKD-$\zeta$ & 86.09& 60.34 & 91.54 & 61.57 & 80.51& 44.04& 83.66& 61.02 \\  \hline
\textbf{FedProtoKD} & \textbf{88.52}& \textbf{61.44}& \textbf{92.27}& \textbf{61.87}& \textbf{81.82}& \textbf{45.58}& \textbf{84.72}& \textbf{61.94}\\ \hline
\end{tabular}
\end{table*}
\subsection{Implementation Details and Hyperparameter}
Unless specified otherwise, all experiments focus on heterogeneous modes. The default framework comprises 10 clients with full participation $\rho = 1$ over $T=100$ communication rounds. Additional evaluations with 20 and 50 clients assess partial participation. Local training spans $ep_c=5$ epochs and server optimization lasts $ep_s=10$ epochs, both utilizing a batch size of 32. The ACTP generator is optimized for $ep_{tsp} = 100$ epochs. Feature dimension is standardized to $k=512$, while Table \ref{table:server_client_acc_client_number_feature_dim} examines dimension variations from 256 to 1024. Regularization weights $\Upsilon$, $\epsilon$, and $\eta$ are fixed at 0.5. The adaptive margin threshold defaults to $\zeta=50$ with sensitivity analysis in Table \ref{tbl:margin_epoch_ACTSP}. Homogeneous model setting results appear only in Tables \ref{table:combined_client_test_accu_homo} and \ref{table:combined_server_test_accu_homo}.
\subsection{Evaluation Criteria}
The system's effectiveness and robustness are assessed by measuring server accuracy, which evaluates the server model's generalization, and Personalized client accuracy, which reflects the client-specific performance. Additionally, the system analyzes Prototype Separation to verify clear class separability in the global prototypes and to quantify the Communication Cost for efficiency assessment.
\begin{figure}[htbp]
\centerline{\includegraphics[width=0.95\linewidth]{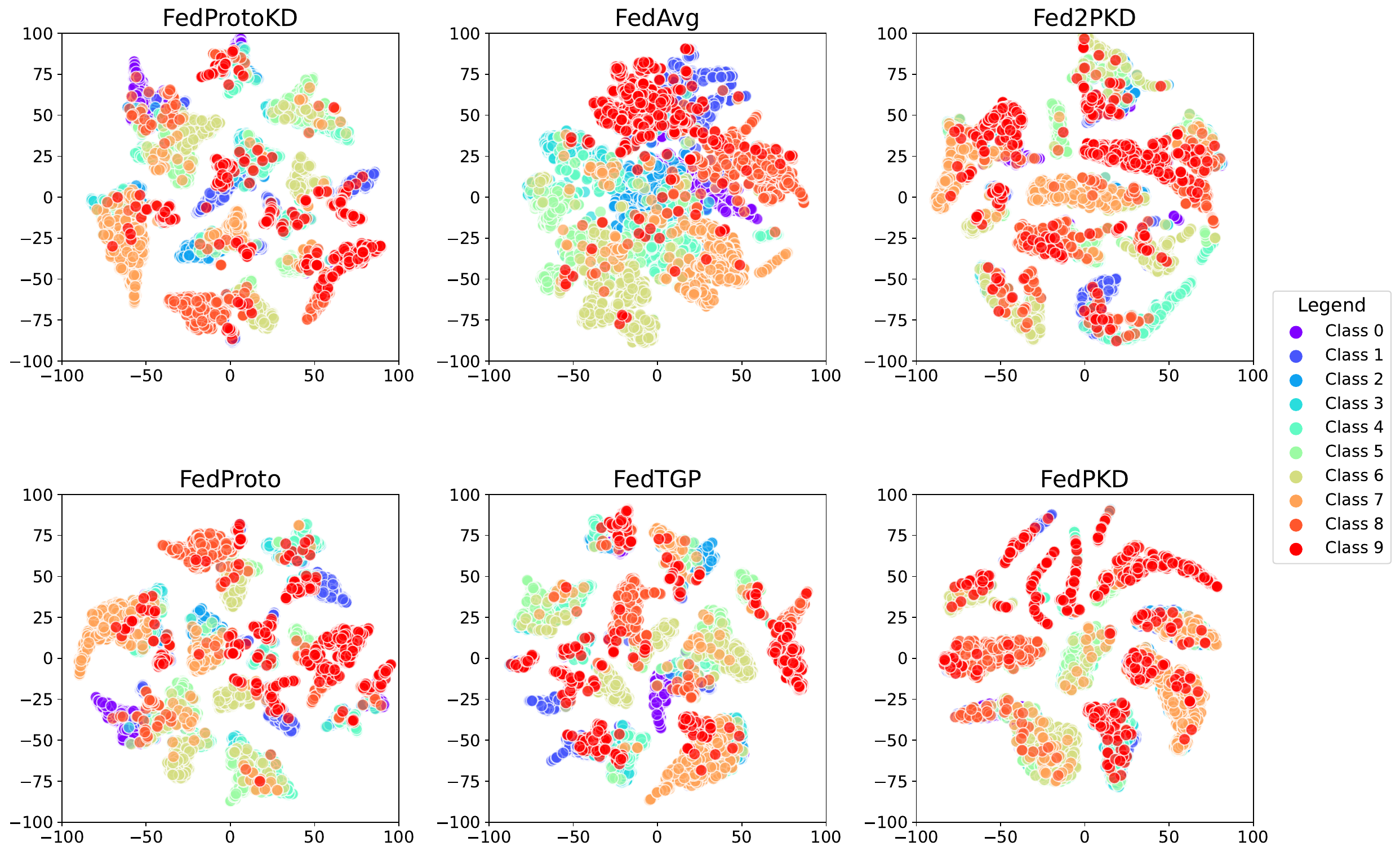}}
\caption{visualization of the feature embeddings in CIFAR-10, $\alpha=0.3$}
\label{fig:margin_shrink_solve_tSNE}
\end{figure}
\section{Experiments Analysis}
\subsection{Enhanced Prototype Margin}
Figure \ref{fig:margin_shrinking} reveals that baseline aggregation suffers from margin shrinkage. Conversely, Figure \ref{fig:margin_shrink_solve} demonstrates that FedProtoKD utilizes the ACTP framework to expand these boundaries, achieving significantly superior inter-class separability compared to competing methods. The maximum prototype margin for FedPKD plateaus around 12, whereas FedTGP and FedProtoKD consistently exceed 30, with our method peaking at 38. This superior geometric regularization is further confirmed by the distinct separation of feature embeddings in Figures \ref{fig:tSNE_class_mean_features_points} and \ref{fig:margin_shrink_solve_tSNE}, demonstrating the efficacy of our approach on CIFAR-100 with $\alpha=0.1$. Table \ref{tbl:margin_epoch_ACTSP} presents ablation results of ACTP for margin growth thresholds $\zeta \in \{10, 50, 100\}$. Unless otherwise specified, the default settings are established with a margin threshold $\zeta = 50$ and 100 ACTP training epochs.
\begin{table}[h]
\centering
\caption{Server and clients Test Accuracy (\%) under Data Heterogeneity on Heterogeneous Model using Tiny-ImageNet}
\label{table:tinyimagenet_combined_client_test_accu_hetero}
\begin{tabular}{|c|cc|cc|} 
\hline
\multicolumn{5}{|c|}{Tiny-ImageNet}  \\ \hline
\multirow{2}{*}{} & \multicolumn{2}{|c|}{$\alpha=0.1$} & \multicolumn{2}{c|}{$\alpha=0.3$} \\ \cline{2-5}
Number of Clients, $K$ & 10 & 20 & 10 & 20 \\ \hline
 & \multicolumn{4}{c|}{Server Accuracy}  \\ \hline
FedPKD &  21.74& 18.45& 16.37& 15.14\\ 
\textbf{FedProtoKD}&  \textbf{22.45}&  \textbf{19.87} &  \textbf{17.64}&  \textbf{15.93}\\ \hline
 & \multicolumn{4}{|c|}{Clients Accuracy}  \\ \hline
FedProto &  24.36& 22.74& 19.46& 16.91\\
FedTGP &  27.26& 24.73& 24.45& 19.25\\
FedPKD &  28.18& 25.03& 25.58& 19.86\\ 
\textbf{FedProtoKD}&  \textbf{29.94} &  \textbf{26.16} &  \textbf{27.71} &  \textbf{21.69} \\ \hline
\end{tabular}
\end{table} 
\begin{table*}[htbp]
\centering
\caption{Server Test Accuracy (\%) on Moderate and Extreme Data Heterogeneity on Homogeneous Models}
\label{table:combined_server_test_accu_homo}
\begin{tabular}{|c|cc|cc|cc|cc|}
\hline
\multirow{4}{*}{} & \multicolumn{4}{c|}{Extreme Data Heterogeneity} & \multicolumn{4}{c|}{Moderate Data Heterogeneity} \\ \cline{2-9}
 & \multicolumn{2}{c|}{$\alpha=0.1$} & \multicolumn{2}{c|}{$k = 3/20$} & \multicolumn{2}{c|}{$\alpha=0.3$} & \multicolumn{2}{c|}{$k = 5/40$} \\ \cline{2-9}
Methods & CIFAR-10 & CIFAR-100 & CIFAR-10 & CIFAR-100 & CIFAR-10 & CIFAR-100 & CIFAR-10 & CIFAR-100 \\ \hline
Fed2PKD & 51.55& 24.63 & 27.45& 16.54& 45.53& 33.45& 35.75& 21.73\\ 
FedPKD &  61.24& 30.98 & 61.02 & 30.20&  71.08&  31.43&  69.41&  31.66\\ \hline
\textbf{FedProtoKD}& \textbf{62.37}&  \textbf{31.74} &  \textbf{62.16} &  \textbf{31.29} & \textbf{70.46}&  \textbf{33.72}&  \textbf{70.93}& \textbf{32.11}\\ \hline
\end{tabular}
\end{table*}
\begin{table*}[htbp]
\centering
\caption{Client Test Accuracy (\%) under Extreme and Moderate Data Heterogeneity on Homogeneous Model}
\label{table:combined_client_test_accu_homo}
\begin{tabular}{|c|cc|cc|cc|cc|}
\hline
\multirow{4}{*}{} & \multicolumn{4}{c|}{Extreme Data Heterogeneity} & \multicolumn{4}{c|}{Moderate Data Heterogeneity} \\ \cline{2-9}
 & \multicolumn{2}{c|}{$\alpha=0.1$} & \multicolumn{2}{c|}{$k = 3/20$} & \multicolumn{2}{c|}{$\alpha=0.3$} & \multicolumn{2}{c|}{$k = 5/40$} \\ \cline{2-9}
Methods & CIFAR-10 & CIFAR-100 & CIFAR-10 & CIFAR-100 & CIFAR-10 & CIFAR-100 & CIFAR-10 & CIFAR-100 \\ \hline
FedProto & 86.28& 60.46& 90.85& 59.99& 79.47& 44.95& 81.49& 59.61\\
FedProx & 86.77& 60.70 & 90.37& 59.60& 80.31& 44.92 & 81.06& 60.17 \\
Fed2PKD & 81.88& 47.44& 76.38& 52.83& 73.84& 37.63& 64.36& 51.30\\
FedTGP & 87.83&  60.14&  91.42&  60.93&  81.14&  44.10&  82.58&  60.82\\
FedPKD & 86.45 & 59.45 & 91.29 & 59.48 & 81.22& 43.83 & 82.07& 59.38 \\ \hline
\textbf{FedProtoKD}&  \textbf{88.41}&  \textbf{62.34}&  \textbf{92.69}&  \textbf{62.36}&  \textbf{82.61}&  \textbf{45.60}&  \textbf{83.73}&         \textbf{62.44}\\ \hline
\end{tabular}
\end{table*} 
\begin{table}[htbp]
\centering
\caption{Server and Client Accuracy (\%) on ACTP Margin Growth Threshold $\zeta$ and Epoch in ACTP for CIFAR-10, $\alpha = 0.1$}
\label{tbl:margin_epoch_ACTSP}
\begin{tabular}{|l|ccc|ccc|}
\hline
\multirow{4}{*}{} &\multicolumn{6}{c|}{Server Accuracy} \\ \cline{2-7}
 & \multicolumn{3}{c|}{Growth Threshold, $\zeta$} & \multicolumn{3}{c|}{Epoch in ACTP} \\ \cline{2-7}
Methods & 10 & 50 & 100 & 50 & 100 & 500 \\ \hline
FedProtoKD-$\zeta$ & 61.27 & 62.11 & 61.73 & 61.33 & 62.11 & 60.19 \\ 
\textbf{FedProtoKD} & \textbf{62.59} & \textbf{63.49} & \textbf{63.18} & \textbf{62.12} & \textbf{63.49} & \textbf{63.43} \\ \hline
& \multicolumn{6}{c|}{Clients Accuracy} \\ \hline
FedTGP & 86.36 & 87.32 & 87.90 & 86.18 & 87.32 & 87.27 \\ 
FedProtoKD-$\zeta$ & 86.79 & 86.09 & 88.42 & 85.53 & 86.09 & 88.15 \\ 
\textbf{FedProtoKD} & \textbf{87.91} & \textbf{88.52} & \textbf{88.48} & \textbf{86.59} & \textbf{88.52} & \textbf{88.62} \\ \hline
\end{tabular}
\end{table}
\subsection{Performance Analysis of Heterogeneous Model Settings}
FedProtoKD is compared against benchmarks supporting model heterogeneity, including FedProto, FedPKD, Fed2PKD, and FedTGP. As shown in Tables \ref{table:combined_server_test_accuracy_hetero_model} and \ref{table:combined_client_test_accuracy_hetero_model}, FedProtoKD demonstrates superior performance across both extreme and moderate heterogeneous data distributions. Regarding server accuracy, FedProtoKD exhibits a significant performance improvement compared to Fed2PKD, with margins of 6.82\% to 34.16\% on CIFAR-10 and CIFAR-100 in extremely heterogeneous scenarios. Under moderate heterogeneity, it retains this advantage, exceeding Fed2PKD by 1.43\% to 26.92\% on these same datasets. This substantial server-side gain is attributed to the trainable prototypes, which provide a stable, architecture-agnostic reference that standard aggregation fails to maintain. This advantage translates robustly to client-side performance as class separability increases. The proposed method outperforms Fed2PKD in extremely heterogeneous cases, yielding improvements of 9.57\% to 15.90\%. On CIFAR-10 and CIFAR-100, FedProtoKD surpasses the competitive FedPKD by up to 3.54\% in client accuracy and exceeds FedTGP by up to 1.84\%. This robustness is further validated on the complex Tiny-ImageNet dataset in Table \ref{table:tinyimagenet_combined_client_test_accu_hetero}, where FedProtoKD achieves 27.71\% client accuracy with $\alpha=0.3$ and 10 clients, surpassing FedPKD by 2.13\% and demonstrating scalability. These results confirm that ACTP mitigates prototype margin shrinkage, ensuring distinct supervision signals despite high heterogeneity.
\subsection{Performance Analysis in Homogeneous Model Settings}
FedProtoKD is evaluated within homogeneous model environments, as Tables \ref{table:combined_server_test_accu_homo} and \ref{table:combined_client_test_accu_homo} demonstrate its superior accuracy. In scenarios of extreme heterogeneity on CIFAR-10, the method improves server accuracy by significant margins relative to Fed2PKD, while moderate heterogeneity yields gains reaching up to 35.18\%. This trend extends to client accuracy on CIFAR-10 and CIFAR-100, where improvements over Fed2PKD remain substantial. Regarding server accuracy, FedProtoKD outperforms FedPKD in most configurations, particularly in Pathological settings. Similarly, client accuracy surpasses FedPKD and FedTGP, with improvements reaching over 3\% on CIFAR-100 in moderate heterogeneous settings . These results highlight the effectiveness of the ACTP framework, where trainable server prototypes robustly adapt to diverse data distributions and mitigate skewing effects.
\begin{table}[htbp]
\centering
\caption{Server and Client Accuracy ($\%$) Achieved under Partial Client Participation ($\mathbf{\rho}$) on Non-IID Data ($\alpha=0.3$). Total Clients $C=20$.}
\label{tbl:partial_cp}
\begin{tabular}{|l|ccc|ccc|}
\hline
\multirow{3}{*}{} & \multicolumn{6}{c|}{Server Accuracy} \\ \cline{2-7}
 & \multicolumn{3}{c|}{CIFAR-10} & \multicolumn{3}{c|}{CIFAR-100} \\ \hline
 Partial Clients, $\rho$& 0.2 & 0.5 & 1.0 & 0.2 & 0.5 & 1.0 \\ \hline
FedPKD & 54.01 & 63.76& 66.64&  14.66&   25.50&  30.09\\ 
\textbf{FedProtoKD} & \textbf{55.65} & \textbf{65.46} & \textbf{69.97}& \textbf{21.10} & \textbf{27.04} & \textbf{30.72}\\ \hline
& \multicolumn{6}{c|}{Clients Accuracy} \\ \hline
FedPKD & 80.32 &  80.83&  79.61&  39.24&  40.53&   41.39\\ 
FedTGP & 80.31 & 80.59 & 77.87& 38.38& 38.98&  38.19 \\
\textbf{FedProtoKD} & \textbf{81.77} & \textbf{81.90} & \textbf{80.72} & \textbf{41.54} & \textbf{41.56} & \textbf{42.19} \\ \hline
\end{tabular}
\end{table}
\begin{table}[htbp]
\centering
\caption{Impact of Client Participation Rate ($\rho$) on Convergence Speed as Global FL Round\# required to achieve Target Accuracy (TA\%)}
\label{tbl:partial_cp_convergence}
\begin{tabular}{|l|cc|cc|}
\hline
\multirow{2}{*}{} & \multicolumn{2}{c|}{Server CIFAR-10} & \multicolumn{2}{c|}{Server CIFAR-100} \\ \cline{2-5} 
Partial Clients, $\rho$ & 0.2 & 1.0 & 0.2 & 1.0 \\ \hline
\multirow{2}{*}{} & TA (50\%) & TA (60\%) & TA (12\%) & TA (25\%) \\ \cline{2-5} 
Methods & Round\# $\downarrow$ & Round\# $\downarrow$ & Round\# $\downarrow$ & Round\# $\downarrow$ \\ \hline
FedPKD & 68 & 45 & 45 & 22 \\ 
\textbf{FedProtoKD} & \textbf{35} & \textbf{13} & \textbf{24} & \textbf{17} \\ \hline
\multicolumn{5}{|c|}{\textbf{Client Performance}} \\ \hline
\multirow{3}{*}{Methods} & TA (80\%) & TA (75\%) & TA (35\%) & TA (35\%) \\ \cline{2-5} 
 & Round\# $\downarrow$ & Round\# $\downarrow$ & Round\# $\downarrow$ & Round\# $\downarrow$ \\ \hline
FedPKD & 53 & 39 & 47 & 18 \\ 
FedTGP & 74 & 42 & 55 & 25 \\ 
\textbf{FedProtoKD} & \textbf{49} & \textbf{31} & \textbf{42} & \textbf{12} \\ \hline
\end{tabular}
\end{table}

\subsection{Robustness of the System}
\subsubsection{Convergence Analysis and Efficiency under Partial Client Participation $\rho$} In Table \ref{tbl:partial_cp_convergence} and \ref{tbl:partial_cp}, FedProtoKD exhibits substantially greater convergence efficiency and robustness compared to FedPKD and FedTGP under Non-IID data conditions, $\alpha=0.3$. It dramatically reduces the Global Rounds (GR\#) needed to achieve target accuracies, particularly when client participation ($\rho$) is $1.0$. For instance, it requires only 13 GR compared to GR 45 for FedPKD, achieving a CIFAR-10 Server target accuracy of 60\%. This acceleration is attributed to efficient dual distillation, which ensures faster convergence and superior accuracy.
\subsubsection{Impact of Client Scalability and Feature Dimensions} 
Table \ref{table:server_client_acc_client_number_feature_dim} evaluates system robustness regarding client scalability and feature dimensionality. The framework exhibits resilience to increased participation as it achieves 69.98\% client accuracy with 50 clients and outperforms FedPKD by 6.42\%. As the number of clients increases, the Non-IID data skewness intensifies, impacting both server and client performance. Regarding feature alignment, a learnable projection layer maps varying architectures to a unified space.
\begin{table}[htbp]
\centering
\caption{Server and client accuracy(\%) on client numbers and prototype dimensions $d$ in CIFAR-10, $\alpha = 0.3$}
\label{table:server_client_acc_client_number_feature_dim}
\begin{tabular}{|l|ccc|ccc|}
\hline
\multirow{3}{*}{} & \multicolumn{6}{c|}{Server Accuracy} \\ \cline{2-7} 
& \multicolumn{3}{c|}{Clients Number, K} & \multicolumn{3}{c|}{Prototype Dimension $d$} \\ \cline{2-7} 
 Methods  &10  & 20        & 50       & 256     & 512       & 1024  \\ \hline 
FedPKD             &  68.52 & 66.64&          61.75& 70.00    & 68.52 & 66.88 \\ 
FedProtoKD-$\zeta$     &   68.61  & 69.87     &          61.69& 69.11  & 68.61  & 69.24 \\
\textbf{FedProtoKD}      &    \textbf{69.45}& \textbf{69.97}& \textbf{64.83}& \textbf{70.34}& \textbf{69.45}& \textbf{70.21}\\ \hline
& \multicolumn{6}{c|}{Client Accuracy} \\ \hline
FedProto      &  79.51&          78.00&            66.82&            79.38&  79.51&                      79.16\\ 
FedTGP      & 80.58&           77.87&             67.06&            80.76&     80.58&                 80.13\\ 
FedPKD      &  79.52&          79.61&            63.56&            79.94&      79.52&               79.61\\ 
FedProtoKD-$\zeta$ &  80.51&          78.47&            66.81&            80.12&   80.51&                   79.79\\  
\textbf{FedProtoKD}  & \textbf{81.82}&           \textbf{80.72}&            \textbf{69.98}&             \textbf{81.38}&    \textbf{ 81.82}&                 \textbf{80.98}\\ \hline
\end{tabular}
\end{table}
\begin{table}[htbp]
\centering
\caption{Client and server accuracy in vary Epoch in client $ep_c$ and server $ep_s$ training in CIFAR-10, $\alpha = 0.3$}
\label{tbl:n_ep_client_server_model}
\begin{tabular}{|c|cc|cc|cc|}
\hline
{$ep_c$, $ep_s$} & \multicolumn{2}{c|}{5, 10} & \multicolumn{2}{c|}{10, 15} & \multicolumn{2}{c|}{15, 20} \\ \hline
                  & Clients & Server & Client & Server & Clients & Server \\ \hline
FedPKD                 &                      79.52&                      68.52&                       85.36&                      59.10&                      88.19&                      60.21\\
FedProtoKD-$\zeta$                 &                      80.51&                      68.61&                      86.21&                      58.28&                      87.46&                      58.36\\ 
\textbf{FedProtoKD}                 &                      \textbf{81.82}&                      \textbf{69.45}&                      \textbf{87.96}&                      \textbf{60.92}&                      \textbf{89.43}&                      \textbf{61.53}\\ \hline
\end{tabular}
\end{table}
\subsubsection{Unlabeled Public Proxy Dataset Size}
\begin{figure}[htbp]
\centerline{\includegraphics[width=0.9\linewidth]{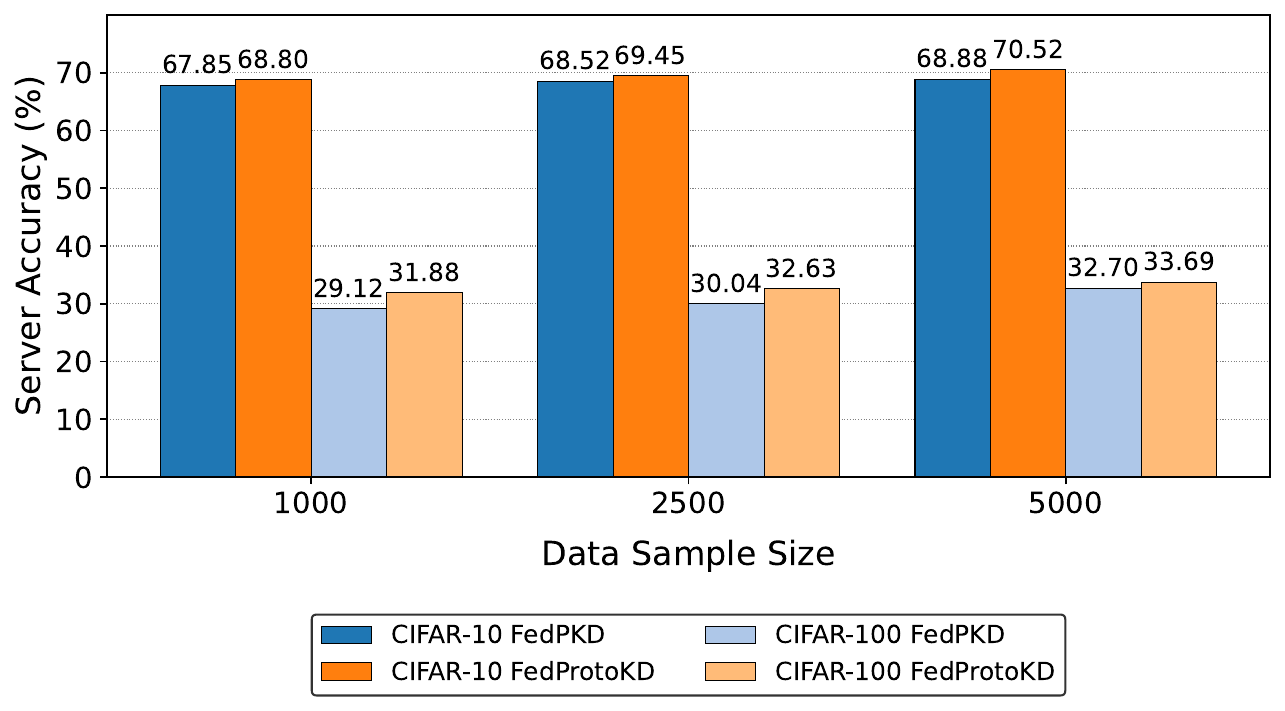}}
\caption{Server accuracy on different public dataset sizes ($\alpha=0.3$)}
\label{fig:public_data_size_vary}
\end{figure}

Figure \ref{fig:public_data_size_vary} analyzes server accuracy relative to public dataset size. FedProtoKD consistently outperforms FedPKD across all sample sizes. Notably, performance peaks at 2,500 samples, where the method achieves 69.45\% accuracy on CIFAR-10 and 32.63\% on CIFAR-100. Extending the dataset to 5,000 samples yields diminishing returns and introduces higher computational overhead. Therefore, a sample size of 2,500 is identified as the optimal operating point to balance distillation efficacy with system efficiency.
\subsubsection{Impact of Training Epochs}
Table \ref{tbl:n_ep_client_server_model} examines the impact of training epoch duration on local and server models. The results indicate an inverse relationship where increasing local epochs improves client accuracy to 89.43\% but degrades server generalization to 61.53\%. This suggests that excessive local training induces client drift and prevents effective global knowledge assimilation. Consequently, a setting of 5 client epochs is selected to preserve global model integrity.
\subsection{Ablation Study}
The individual contributions of the proposed components are evaluated via an ablation study in Table \ref{tbl:ablation_experiments} which isolates the ACTP and Public Data Sample Importance $\mathcal{I}$. Two variants are analyzed. FedProtoKD with fixed $\zeta$ replaces the adaptive margin while the second variant excludes the sample importance mechanism. Results confirm that both components are critical. Integrating sample importance increases CIFAR10 accuracy from 61.96\% to 63.49\% at $\alpha=0.1$ proving that importance weights effectively filter noise. Furthermore, the adaptive margin outperforms the fixed approach by raising CIFAR100 accuracy from 30.23\% to 31.80\%. Therefore, FedProtoKD achieves higher performance across all settings, validating the use of ACTP and selective sample distillation in handling extreme heterogeneity.
\begin{table}[htbp]
\centering
\caption{Impact of ACTP and Sample Importance on Server Accuracy}
\label{tbl:ablation_experiments}
\begin{tabular}{|c|c|ccc|}
\hline
Dataset & $\alpha$ & FedProtoKD-$\zeta$ & W/O \(\mathcal{I}\) & \textbf{FedProtoKD} \\ \hline
CIFAR-10  & 0.1  & 61.14 & 61.96  & \textbf{63.49} \\  
CIFAR-100 & 0.1 & 30.23 & 30.41  & \textbf{31.80} \\ \hline
&$k$ &  &  &  \\ \hline
CIFAR-10  & 3  &      58.37&       58.15&      \textbf{61.97}\\  
CIFAR-100 & 20 &      29.12&       29.63&      \textbf{30.88}\\ \hline
\end{tabular}
\end{table}
\section{Related Work}
\label{gen_inst}
\subsubsection*{Heterogeneous Federated Learning}  Statistical heterogeneity (non-IID data) is a major challenge in Federated Learning. Methods such as FedProx \cite{TLiFedProx2020}, regularization-based personalization \cite{t2020personalized}, model mixtures \cite{ mansour2020three}, client clustering \cite{sattler2020clustered}, and meta-learning \cite{fallah2020personalized} mitigate local overfitting and improve stability. However, these approaches assume homogeneous model architectures and are insufficient when clients use different models. Heterogeneous Federated Learning (HFL) addresses both statistical and architectural heterogeneity. Some HFL methods allow clients to sample submodels from a global model, while others share only upper layers in LG-FedAvg \cite{liang2020think}, FedGen \cite{zhu2021data}. Yet, aggregating shared layers can still yield suboptimal results under severe statistical heterogeneity.
\subsubsection*{Federated Ensemble Distillation} Knowledge Distillation is a powerful tool for handling model heterogeneity in FL. FedDF \cite{lin2020ensemble} ensembles local models as a teacher to train the global student model, improving robustness under non-IID data. FedBE \cite{chen2020fedbe} models local parameter distributions via Bayesian inference and samples additional teachers for stronger ensemble distillation. FedPKD \cite{lyu2023prototype} selects high-quality proxy samples using class-wise prototypes but relies on public data and does not account for the quantitative effects of domain drift.
\subsubsection*{Prototype Learning} Prototypes, defined as mean feature representations, are widely used in tasks such as image classification. FedProto \cite{tan2022fedproto} aggregates client prototypes for regularization without training a global model. FedProc \cite{mu2023fedproc} introduces prototypical contrastive learning but does not support heterogeneous models since it exchanges parameters. Fed2PKD \cite{xie2024fed2pkd} combines prototypical contrastive distillation with semi-supervised global distillation. FedPKD \cite{lyu2023prototype} enhances client–server performance and reduces communication via prototype learning, while FedTGP \cite{JZhang2024} introduces trainable global prototypes with maximum prototype margin.
\section{Conclusion}
This paper presents FedProtoKD, a novel Heterogeneous Federated Learning framework designed to mitigate the prototype margin shrinkage problem through class-wise adaptive margins. By integrating a contrastive learning-based generator with a public sample importance mechanism, the proposed method facilitates robust knowledge distillation across both classification and feature learning dimensions. This dual-level alignment effectively manages extreme model and data heterogeneity while preserving privacy. Extensive experiments demonstrate that FedProtoKD consistently outperforms state-of-the-art benchmarks by enhancing class separability and distillation efficiency. Future research will extend this framework to larger-scale datasets and diverse foundation models to further evaluate system robustness.

\end{document}